\documentclass[10pt,twocolumn,letterpaper]{article}

\usepackage[pagenumbers]{wacv} 

\usepackage{graphicx}
\usepackage{amsmath}
\usepackage{amssymb}
\usepackage{booktabs}
\usepackage[dvipsnames]{xcolor}
\usepackage[accsupp]{axessibility} 

\usepackage[pagebackref,breaklinks,colorlinks]{hyperref}

\usepackage[capitalize]{cleveref}
\crefname{section}{Sec.}{Secs.}
\Crefname{section}{Section}{Sections}
\Crefname{table}{Table}{Tables}
\crefname{table}{Tab.}{Tabs.}

\def\wacvPaperID{2481} 
\def\confName{WACV}
\def\confYear{2025}

\newcommand{\ours}{V-MIND}
\newcommand{\mparagraph}[1]{\vspace{0.3em}\noindent\textbf{#1}}
\newcommand{\mvec}[1]{\mathbf{#1}}
\newcommand{\mmat}[1]{\mathbf{#1}}
\definecolor{green}{rgb}{0.0, 0.5, 0.0}
\definecolor{red}{rgb}{0.8, 0.0, 0.0}

\begin{document}

\title{{\ours}: Building Versatile Monocular Indoor 3D Detector with Diverse 2D Annotations}


\author{
    Jin-Cheng Jhang$^{1}$\thanks{Equal contribution}
    \quad
    Tao Tu$^{2,3}$\footnotemark[1] \quad
    Fu-En Wang$^{3}$ \quad
    Ke Zhang$^{3}$ \quad
    Min Sun$^{1,3}$\thanks{Equal corresponding authors} \quad
    Cheng-Hao Kuo$^{3}$\footnotemark[2] \quad \\
    $^{1}$National Tsing Hua University \quad 
    $^{2}$Cornell University \quad 
    $^{3}$Amazon\\
    {\tt\small frank890725@gapp.nthu.edu.tw \quad
    tt582@cornell.edu}
    \\
    {\tt\small \{fuenwang, kezha, minnsun, chkuo\}@amazon.com
    }
}
\maketitle


\begin{abstract}
The field of indoor monocular 3D object detection is gaining significant attention, fueled by the increasing demand in VR/AR and robotic applications.
However, its advancement is impeded by the limited availability and diversity of 3D training data, owing to the labor-intensive nature of 3D data collection and annotation processes.
In this paper, we present \textbf{\ours} (\textbf{V}ersatile \textbf{M}onocular \textbf{IN}door \textbf{D}etector), which enhances the performance of indoor 3D detectors across a diverse set of object classes by harnessing publicly available large-scale 2D datasets.
%
By leveraging well-established monocular depth estimation techniques and camera intrinsic predictors, we can generate 3D training data by converting large-scale 2D images into 3D point clouds and subsequently deriving pseudo 3D bounding boxes.
To mitigate distance errors inherent in the converted point clouds, we introduce a novel 3D self-calibration loss for refining the pseudo 3D bounding boxes during training. 
Additionally, we propose a novel ambiguity loss to address the ambiguity that arises when introducing new classes from 2D datasets.
Finally, through joint training with existing 3D datasets and pseudo 3D bounding boxes derived from 2D datasets, {\ours} achieves state-of-the-art object detection performance across a wide range of classes on the Omni3D indoor dataset.
%



\end{abstract}
\section{Introduction}
\label{section:intro}

Given a single 2D image, humans possess the remarkable ability to recognize a diverse set of object classes, effortlessly estimating their locations, sizes, and poses in 3D space.
This task is referred to as Monocular 3D Object Detection (M3OD).
In comparison to 2D object detection, M3OD presents greater challenges as it requires estimating the distance and size of objects from a single 2D visual input, which is inherently prone to ambiguity.
With the surge in applications of VR/AR and robotics, 3D object detection has gained significant attention.
%
%
However, state-of-the-art 3D object detectors support only a limited number of object classes due to the scarcity of training data with 3D annotations.
To unlock the potential of large-vocabulary 3D object detectors, we study indoor environments in this work, where a much broader array of objects---encompassing diverse shapes, sizes, and categories---presents a unique challenge compared to the typical outdoor street-view settings.
This is evident from the fact that the number of object classes in Omni3D's indoor environments is six times greater than that in outdoor environments~\cite{omni3d}.



2D object detection has experienced significant progress in the past few years. Firstly, crowdsourcing has scaled up manual 2D annotation to a few thousand classes~\cite{gupta2019lvis}. Moreover, with the advance of shared image and text embedding (e.g., CLIP~\cite{radford2021learning}), generalizing 2D object detection to an open-set of classes has become one of the hottest topics in computer vision (referred to as open-vocabulary detection). Recently, several methods~\cite{lu2023open,cao2023coda} generalize the open-vocabulary capability to 3D detection. However, these methods require that 3D structures be sensed or reconstructed from multiple views. The 3D structure is the key to predicting the 3D boxes, and the 2D images are the key to predicting the classes. We contend that a monocular solution holds broader applicability, as it circumvents the need for expensive 3D sensors and can be applied to each frame in an online manner.

\begin{figure*}[th]
    \centering
    \includegraphics[width=0.97\textwidth]{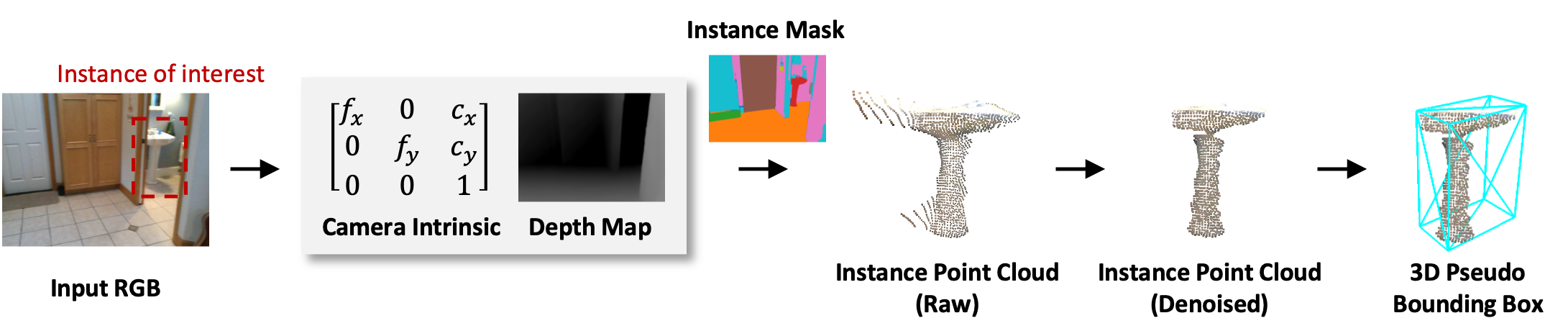}
    \vspace{-.8em}
    \caption{\textbf{Augmented 2D-to-3D dataset generation.}
    Given a 2D image dataset with diverse object classes, we first use a metric-based depth estimator to predict the depth map and an intrinsic predictor to derive the camera's intrinsic parameters.
    We then compute and denoise the object point cloud for each instance and generate a 3D bounding box (\Cref{method:3d-data-generation}) that tightly encloses the point cloud.
    }
    \label{fig:3d-data-generation}
\end{figure*}

We recognize that the primary bottleneck in M3OD lies in the scarcity of annotated 3D training data.
One straightforward and effective solution entails lifting the 2D training data into 3D by capitalizing on advancements in monocular 3D reconstruction. Our proposed lifting process is shown in \Cref{fig:3d-data-generation} and unfolds as follows: firstly, we employ a state-of-the-art monocular \emph{metric} depth estimator~\cite{bhat2023zoedepth} to predict the depth of a 2D image. Subsequently, we utilize a cutting-edge predictor to predict the image's intrinsic properties~\cite{zhu2024tamecamera}. Next, given an object segment, we lift 3D point clouds from the object's surface into the camera coordinate system, followed by the application of a rapid filtering method to eliminate spurious points. Finally, we compute a precise 3D bounding box encapsulating the object, filtering out objects with insufficient supporting point clouds based on their classes.


The lifted 3D boxes on a large-scale 2D dataset like LVIS~\cite{gupta2019lvis} can outnumber those in real-world 3D datasets such as SUN-RGBD~\cite{SUNRGBD} by nearly 10 times.
We jointly train the proposed {\ours} using the manually labeled 3D and lifted 2D data. To mitigate the remaining distance error, we propose a 3D \emph{self-calibration loss} for the lifted 3D boxes.
These errors are caused by errors in metric depth and camera intrinsic prediction.
%
%
Furthermore, in addition to adding more examples for existing classes, which we call the original classes, incorporating diverse large-scale 2D datasets can also introduce new classes.
However, these new class objects may exist in the original 3D datasets but lack ground truth annotations. In such cases, the objects might be incorrectly assigned as background, leading to ambiguity during training. Therefore, instead of solely relying on cross-entropy loss for classification, we introduce a novel \textit{ambiguity loss} to address this ambiguous problem.
We conduct experiments on the indoor Omni3D~\cite{omni3d} dataset (Omni3D\textsubscript{IN}) and show that significant accuracy improvement can be achieved by training with lifted 2D data. Moreover, we train the model using a CLIP embedding classifier and show further improvement in long-tailed classes.

Our main contributions can be summarized as follows:
\begin{itemize}
\setlength\itemsep{.1em}
    \item We present a simple yet effective method for lifting large-scale 2D labeled datasets into training data for 3D object detection.
    \item We introduce {\ours}, a monocular 3D object detector that incorporates a novel self-calibration loss for lifted 3D boxes to reduce distance errors and a novel ambiguity loss to resolve ambiguities when training with new classes, achieving state-of-the-art detection accuracy on the Omni3D\textsubscript{IN}. 
    \item Our proposed method will continuously benefit from more 2D labeled data and improvement of monocular 3D reconstruction methods.
\end{itemize}


\section{Related Work}
\subsection{2D Object Detection}
There are two kinds of mainstream approaches for 2D object detection: 1) Two-stage and 2) Single-stage approaches.
\cite{rcnn,fastrcnn,fasterrcnn} are pioneer works for two-stage approaches, in which the first stage adopts a region proposal network (RPN) to produce class-agnostic bounding boxes, and the second stage uses a classification network to infer object categories. Inspired by \cite{fasterrcnn}, \cite{maskrcnn} incorporated an additional segmentation head to infer object mask to further improve detection performance. For the single-stage approach, \cite{yolo} proposed to adopt a single convolutional network to extract anchor features, followed by a fully connected network to infer bounding box coordinates. To improve the performance of small objects, \cite{ssd} proposed using multi-scale feature maps to leverage both large and small objects in a single-stage scheme. \cite{cornernet,centernet} proposed single-stage anchor-free object detection frameworks, which directly use centerness heatmaps and offsets to improve both detection performance and efficiency.
In recent years, transformer-based detectors have gained popularity due to their better scalability and improved accuracy on challenging benchmark datasets. DETR~\cite{DETR} is the first transformer-based detector with a one-stage design and without the need for anchor and NMS. Deformable DETR~\cite{deformabledetr} proposed a two-stage method where region proposals are generated at the first stage. The generated region proposals will be fed into the decoder as object queries for further refinement. More transformer-based detectors are studied and compared in Shehzadi et al.~\cite{shehzadi2023object}.
In this work, we also select a two-stage design for better accuracy.




\subsection{Monocular 3D Object Detection}
In this work, we focus on the indoor M3OD, a more general scenario compared to multi-view detection~\cite{imvoxelnet, tu2023imgeonet}.
On the method side, Huang et al.~\cite{huang2018cooperative} focus on predicting 3D oriented bounding boxes for indoor objects, while Factored3D~\cite{factored3dTulsiani17} and 3D-RelNet~\cite{Kulkarni-2019-121566} concurrently predict object voxel shapes. Total3D~\cite{Total3D} extends its predictions to encompass 3D boxes and meshes through additional training on datasets featuring annotated 3D shapes. ImVoxelNet~\cite{Rukhovich_2022_WACV} introduces a foundational framework for processing volumes of 3D voxels.
On the dataset side, Omni3D~\cite{omni3d} is a recent attempt to scale the annotated 3D training data by combining SUN-RGBD~\cite{SUNRGBD}, ARKitScenes~\cite{dehghan2021arkitscenes}, and Hypersim~\cite{hypersim} for indoor objects. In Omni3D, a Cube R-CNN detector is trained on three datasets and achieves consistent performance gain on three datasets in ``10" intersecting categories. This shows the positive impact of scaling up training data but also highlights the limited class annotated in these datasets. In this work, we leverage 2D datasets with a large number of classes annotated. Hence, a performance gain can be observed in more long-tail classes.
Recently, UniMode~\cite{unimode} introduced a BEV-based detector. Working in the BEV space, it avoids the error introduced during the conversion from the 2D camera plane to the 3D physical space. On the Omni3D~\cite{omni3d} dataset, it outperforms Cube R-CNN. However, since their code was not released at the time of our submission, we could not make a direct comparison. Nevertheless, we believe our proposed approach can also be applied to their method for further improvement.


\subsection{Large and Open Vocabulary Object Detection}
In \textbf{2D object detection}, \cite{gupta2019lvis,Redmon_2017_CVPR,8578217,Yang_2019_ICCV} require detecting over 1000 classes (referred to as Large-Vocabulary). Many existing works focus on addressing the long-tail problem \cite{Feng_2021_ICCV,li2020overcoming,NEURIPS2021_14ad095e,Zhang_2021_CVPR}. Equalization losses~\cite{Tan_2021_CVPR,Tan_2020_CVPR} and SeeSaw loss~\cite{Wang_2021_CVPR} reweight the per-class loss by balancing the gradients~\cite{Tan_2021_CVPR} or the number of samples~\cite{Wang_2021_CVPR}. Yang et al.~\cite{Yang_2019_ICCV} detect 11K classes using a label hierarchy. Detic~\cite{Detic} expands the training data to include additional image-labeled data. 
Open-vocabulary object detection (also known as zero-shot detection) is the task of detecting novel classes for which no training labels are provided~\cite{gu2022openvocabulary, RahmanAAAI20,ZareianCVPR21}. A common solution is to replace the classifier with a pre-trained vision-language embedding~\cite{RahmanAAAI20,ZareianCVPR21}, allowing the detector to utilize an open-vocabulary classifier and perform open-vocabulary detection directly. 
Note that many works like Detic~\cite{Detic} show that using pre-trained vision-language embedding as classifiers can also improve the large-vocabulary classes seen in training.
In this work, we focus on large-vocabulary for Monocular 3D object detection. Hence, we also utilize pre-trained vision-language embedding to enhance the accuracy of large-vocabulary.
In \textbf{3D object detection}, most methods utilize point cloud at inference time to address the challenge of localization in 3D. This is because 3D datasets are much smaller than 2D datasets, as noted in~\cite{gao2023dqs3d}. Consequently, image-based 3D region proposals are much less accurate than their 2D counterpart. 
Recently, OV-3DET~\cite{lu2023openvocabulary} proposed a 3D point cloud-based 3D object detector learning to align point cloud-based feature with pre-trained CLIP~\cite{radford2021learning} feature space. OV-3DET leverages a pre-trained OV-2D detection model~\cite{Detic} to generate 3D pseudo boxes so that OV-3DET can be trained without human annotation.
CoDA~\cite{cao2023coda} tackles 3D OV detection in a different setting. It assumes a set of base classes are available with ground truth 3D boxes. Then, an iterative novel object discovery and model enhancement procedure is proposed.
However, all these methods rely on 3D point clouds to be available during inference, whereas we focus on monocular image-based 3D object detectors which is more generally applicable.

\label{section:relatedwork}
\section{Methodologies}
%
To address the scarcity of annotated 3D training data, we generate pseudo 3D training data by lifting existing internet-scale 2D data into 3D space (\Cref{method:3d-data-generation}), utilizing the state-of-the-art metric depth estimator and monocular camera intrinsic predictor. The lifted 3D data is further cleaned and processed automatically into 3D boxes (referred to as generated 3D data).
We then train {\ours} using both annotated 3D data and our generated 3D data. To fully exploit the rich semantic information in the 2D data, we perform object classification in a pre-trained vision-language space~\cite{radford2021learning}, where similar feature vectors represent similar visual concepts (\Cref{method:3d-detector}).
Additionally, to mitigate potential errors in the 3D lifting process, we introduce a 3D self-calibration loss that enables {\ours} to self-calibrate and accurately locate objects (\Cref{method:self-calib}).
Finally, we elaborate on the ambiguity issue when incorporating new classes from the lifted 2D data and introduce the proposed ambiguity loss to address this issue (\Cref{method:ambiguity_loss}).

\begin{figure*}[h]
    \centering
    \includegraphics[width=\textwidth]{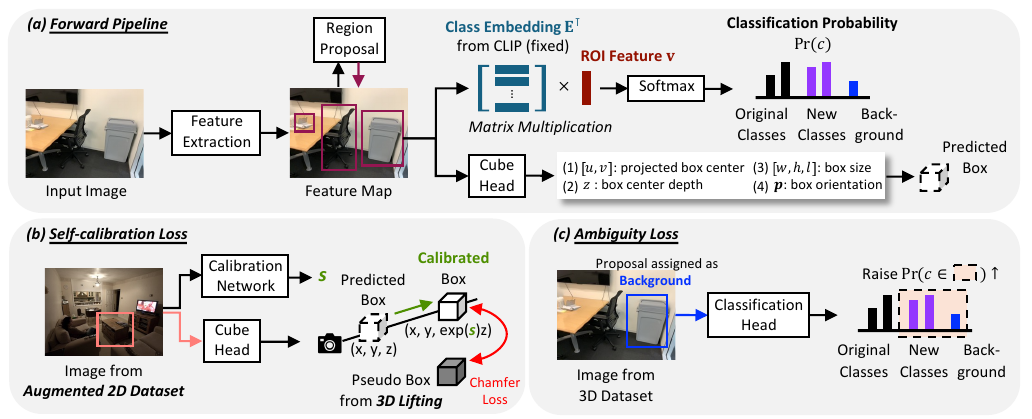}
    \caption{\textbf{Overview of the proposed {\ours}.}
    (a) Given an input image, a backbone model extracts features, which are then used to generate region proposals. Each proposal is classified in a pre-trained vision-language space, with a 3D bounding box predicted by the cube head (\Cref{method:3d-detector}). Leveraging an augmented 2D-to-3D dataset, our model can detect not only the original object classes but also previously unseen ones, \ie, the new classes.
    (b) Since pseudo 3D boxes from the augmented dataset may contain lifting errors, we propose calibrating the predicted 3D boxes before computing the Chamfer loss. This adjustment helps prevent performance degradation caused by lifting errors (\Cref{method:self-calib}).
    (c) In the 3D training dataset, objects from new classes may be treated as background during proposal assignment (\Cref{method:ambiguity_loss}). To address this, we maximize the probability of the combined group of new classes and the background class, rather than just the background.
    }
    \label{fig:architecture}
\end{figure*}

\subsection{Large-scale 3D Training Data Generation}
\label{method:3d-data-generation}
Given large-scale 2D images $I$ and diverse instance masks $M$, we generate semantically rich 3D training data by lifting all annotated instances in the images to 3D (\Cref{fig:3d-data-generation}).
We use an off-the-shelf monocular metric depth estimator~\cite{bhat2023zoedepth} and camera intrinsic predictor~\cite{zhu2024tamecamera} to generate depth maps $Z$ and obtain the camera intrinsic parameters $K$.
Then the 3D point cloud of each instance can be derived by
\begin{equation}
\label{eq:backproj}
    \begin{bmatrix}
    x \\
    y \\
    z
    \end{bmatrix}
    = ZK^{-1}
    \begin{bmatrix}
    u \\
    v \\
    1
    \end{bmatrix},
\end{equation}
where $u$ and $v$ are the 2D pixel coordinate of a given instance.

\mparagraph{Cleaning Lifted 3D Data.}
Due to suboptimal depth maps and inaccuracies in the camera intrinsic parameters, the outliers in the generated 3D point clouds degrade the quality of our 3D training data.
Therefore, we remove the outliers that either have few neighboring points within a given radius or are distant from their nearest neighbors.

\mparagraph{3D Boxes Generation.}
As our goal is 3D detection, we use an off-the-shelf minimum bounding box algorithm to generate oriented 3D bounding boxes that tightly enclose the instance point clouds.
Besides, we filter out those 3D bounding boxes with very few points based on our proposed per-category point threshold strategy. The point threshold of each category is calculated according to the average number of object points in ScanNet200~\cite{rozenberszki2022scannet200}, where the point cloud of each object can be obtained by the ground-truth semantic segmentation and depth maps.
%
%
%
Furthermore, to reduce the noises in the ground-truth semantic segmentation masks, we refine the masks using a state-of-the-art segmentation model~\cite{qi2023cropformer}.
For the categories not included in ScanNet200, we map these categories to their closest ones in ScanNet200 using the cosine distances between their CLIP~\cite{radford2021learning} embeddings and use the object points of mapped categories as final thresholds.
%


\subsection{Monocular 3D Object Detector}
\label{method:3d-detector}
\mparagraph{Preliminary of Cube R-CNN.}
Our monocular 3D detector builds on Cube R-CNN~\cite{omni3d}, an end-to-end monocular 3D object detector that predicts 3D boxes in virtual depth space, allowing the detector to handle images with varying camera intrinsic parameters during training.
Cube R-CNN comprises a region proposal network (RPN) for generating 2D region proposals, a 2D box head for detecting object class and predicting 2D bounding box's 2D center $[x_\text{2D}, y_\text{2D}]$ and size $[w_\text{2D}, h_\text{2D}]$, and a cube head for outputting a 3D cuboid corresponding to each region proposal.
Four components parameterize the output 3D cuboid of Cube R-CNN:
(1) $[u, v]$, the projected box center on the 2D image relative to the 2D region of interest. (2) $z$, the depth of the projected box center. (3) $[w, h, l]$, the lengths of each side. (4) $\mvec{p}$, the 6D rotation representation~\cite{zhou2019continuity} for the box orientation.
The coordinate of the predicted 3D box center, $\mvec{x}$, is then $[\frac{z}{f_\text{x}}(x_\text{2D} + u\cdot w_\text{2D} - p_\text{x}), \frac{z}{f_\text{y}}(y_\text{2D} + v\cdot h_\text{2D} - p_\text{y}), z]$, where $(p_\text{x}, p_\text{y})$ is the principal point offset.

\mparagraph{Large-Vocabulary Detection.}
Inherit from the rich semantic classes in internet-scale 2D data, our pseudo 3D data contains various object classes.
The large-scale object classes pose a challenge to conventional detection models, as each unique category name is treated as a unique class. None of the knowledge between category names is used during training.
For instance, "toilet tissue" and "toilet paper" are treated as different classes as a result the model is trained to distinguish between "toilet tissue" and "toilet paper".
%
%
To fully leverage such semantically rich data and address the aforementioned issue, we classify objects in a pre-trained vision-language space~\cite{radford2021learning}.
As similar visual concepts are represented by similar features in this pre-trained vision-language space, regardless of the differences in their textual descriptions, our detector can concentrate on learning semantic concepts rather than discrete categories.
Specifically, we generate class embeddings for the classes in our diverse dataset using the pre-trained text encoder.
As shown in \Cref{fig:architecture}~(a), we then classify a region proposal by computing the classification probability $\Pr(c)$ through the cosine similarity between the predicted region feature $\mvec{v}$ and the class embeddings $\mmat{E}$ as
\begin{equation}
\label{eq:clip-classification}
\Pr(c_i)=\mathrm{softmax}\left(\frac{\mmat{E}_{i}^\intercal\mvec{v}}{\left \lVert\mmat{E}_i\right\rVert_2\left \lVert\mvec{v}\right\rVert_2}\right),
\end{equation}
where $c_i$ is the $i$-th object class.

\subsection{Self-calibration Loss}
\label{method:self-calib}
Since the generated 3D pseudo bounding boxes have lifting errors resulting from the imperfect depth maps and camera intrinsic parameters, we allow the detector to self-calibrate the output box before computing Chamfer loss by a predicted scale from a learnable calibration network, which is only applied for the pseudo 3D data during training (\Cref{fig:architecture} (b)).
Specifically, the output cuboid is computed as
\begin{equation}
\label{eq:3d-cuboid}
    B_{\text{3D}} = \mmat{R}(\mvec{p})\,\mathrm{diag}(w,h,l)\,B_{\text{unit}} + \exp(s)\mvec{x},
\end{equation}
where $\mmat{R}(\mvec{p})$ is the rotation matrix representing the box orientation~\cite{omni3d}, $\mathrm{diag}(w,h,l)$ is a diagonal matrix where each diagonal entry corresponds to a side length of the box, column-major matrix $B_{\text{unit}}$ are the eight corners of an axis-aligned unit box centered at the origin, $\mvec{x}$ is the predicted 3D box center, and $\exp(s)$ is the self-calibrated scale for each image.
%
To prevent the model from overfitting, we add a regularization term for the self-calibrated scale:
\begin{equation}
\label{eq:self-calibration-regularization}
    \mathcal{L}_\text{reg} = \left\lVert s \right\rVert _1.
\end{equation}
On the other hand, to prevent the lifting errors from pseudo 3D data affecting the 2D visual feature extraction for the original classes, we detach the input features before passing them to the calibration network. 
The network architecture is described in the supplementary material.

\subsection{Ambiguity Loss}
\label{method:ambiguity_loss}
In addition to adding more examples for existing classes, incorporating diverse large-scale 2D datasets will also introduce new classes.
However, when introducing new classes from the augmented 2D-to-3D dataset (LVIS~\cite{gupta2019lvis}), it is inevitable to encounter an ambiguous situation where new objects may exist in the original 3D dataset (Omni3D~\cite{omni3d}) but are not included in the ground truth annotations.
Consequently, these objects will be wrongly assigned as background during training.
In this scenario, applying the standard cross-entropy loss will result in incorrect penalties even when the model successfully predicts these new objects.

\begin{figure}[h]
    \centering
    \includegraphics[width=0.42\textwidth]{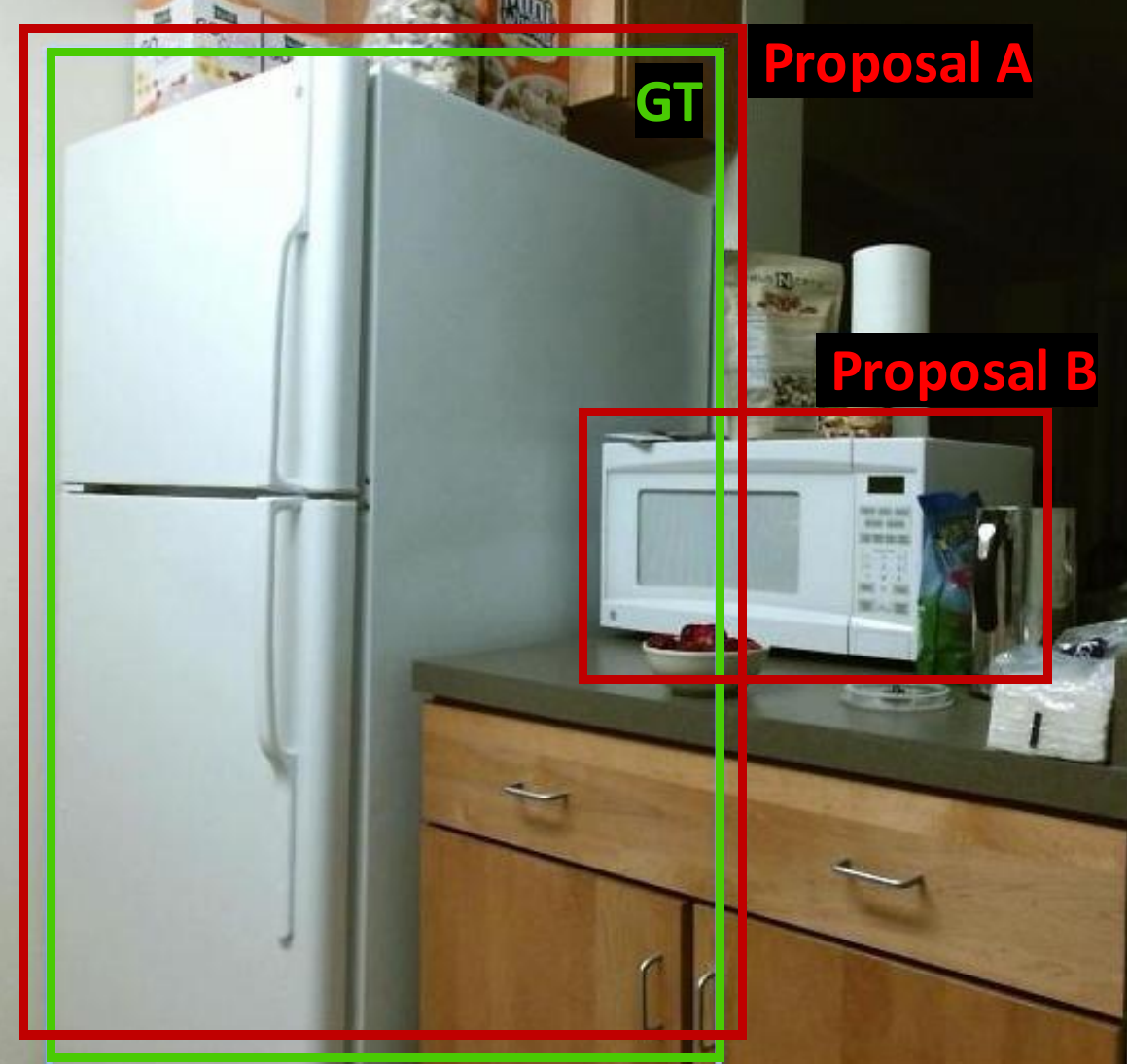}
    \caption{\textbf{Example to illustrate the ambiguity issue.}
    Data from the 3D dataset with a limited vocabulary treats proposals enclosing new-class objects as background (e.g., proposal B in the figure), which interferes with the training of the 3D detector and hinders its accuracy in identifying new classes.
    %
    %
    }
    \label{fig:ambiguous_loss_example}
\end{figure}

We use \Cref{fig:ambiguous_loss_example}, an image from the Omni3D dataset, to further illustrate the ambiguity issue. 
In this image, a refrigerator and a microwave need to be detected, belonging to the original and new classes, respectively.
Since the image is from the Omni3D dataset, only the original classes have the ground truth labels during training, as indicated by the green box in \Cref{fig:ambiguous_loss_example}.
In this case, the predicted proposal A is assigned to a correct label, allowing the classification loss to be computed using cross-entropy loss as usual.
However, the predicted proposal B corresponds to a new class (microwave) that lacks annotation during training.
As a result, proposal B will be incorrectly assigned to the background class.
If the model successfully classifies this new object (microwave), the cross-entropy loss will mistakenly penalize this prediction, which can severely hinder the model's ability to learn accurate classifications for new classes.

To this end, we propose the \textit{ambiguity loss} $\mathcal{L}_\text{amb}$.
Since a proposal assigned as background in the 3D dataset (Omni3D) could belong to either a new class or the background, we encourage the model to increase the probability of both by minimizing the following:
\begin{equation}
\label{eq:disambiguous_loss}
    \mathcal{L}_\text{amb} = -\log\Bigl(\sum_{c\in \Omega_\text{new}\cup\Omega_\text{bkg}}\Pr(c)\Bigr),
\end{equation}
where $\Omega_\text{new}$ and $\Omega_\text{bkg}$ represent the sets of new classes and the background class, respectively.
This objective helps prevent incorrect penalties caused by the ambiguity issue.
%
%
%
%



\section{Experiments}
\label{section:experiments}
%

\subsection{Experiment Settings}
\label{sec.expsetting}
\mparagraph{Dataset.}
We generate our pseudo 3D training data from LVIS~\cite{gupta2019lvis}, a large-scale 2D image dataset providing high-quality instance masks.
LVIS contains 118,287 images and 1200+ object classes.
For the manually annotated 3D data, we use the Omni3D~\cite{omni3d} dataset, specifically the indoor subset ($\text{Omni3D}_{\text{IN}}$), which contains 125,528 training images and 84 object classes from SUN-RGBD~\cite{SUNRGBD}, ARKitScenes~\cite{dehghan2021arkitscenes}, and Hypersim~\cite{hypersim} datasets.

\mparagraph{Original class and New class.}
Although the $\text{Omni3D}_{\text{IN}}$ contains 84 classes, only 38 classes were selected for training and evaluation in their paper. Therefore, we define the same 38 classes as the \textit{original class}. Among the remaining 46 classes, we also select 38 classes as the \textit{new class}, excluding those with relatively few instances in the LVIS dataset. These new class annotations in the $\text{Omni3D}_{\text{IN}}$ will not be included in the training data, adhering to the setting in their paper, and will only be used for evaluation.
For the LVIS dataset, we include and merge the classes that align with the definitions of both the \textit{original class} and \textit{new class}, based on the cosine similarity of their text embeddings extracted from the pre-trained text encoder~\cite{radford2021learning}.
We provide detailed statistics of the training data in the supplementary material.
In total, we have 76 classes, doubling the vocabulary size compared to the original $\text{Omni3D}_{\text{IN}}$ setting.

\mparagraph{Metric.}
We report the overall mean average precision for both 2D and 3D object detection accuracy (AP\textsubscript{2D} and AP\textsubscript{3D}), and also separately report the ones for the \textit{original class} and the \textit{new class} 
($\text{AP}_\text{2D}^\text{original}$ and $\text{AP}_\text{2D}^\text{new}$).
%
Following Omni3D benchmark~\cite{omni3d}, we compute AP across intersection over union (IoU) thresholds ranging from $0.05$ to $0.50$.
Here, the IoU threshold denotes the minimum IoU required to consider a match as positive.

\subsection{Results}
\label{sec.expresult}
\begin{figure}[h]
    \centering
    \includegraphics[width=0.47\textwidth]{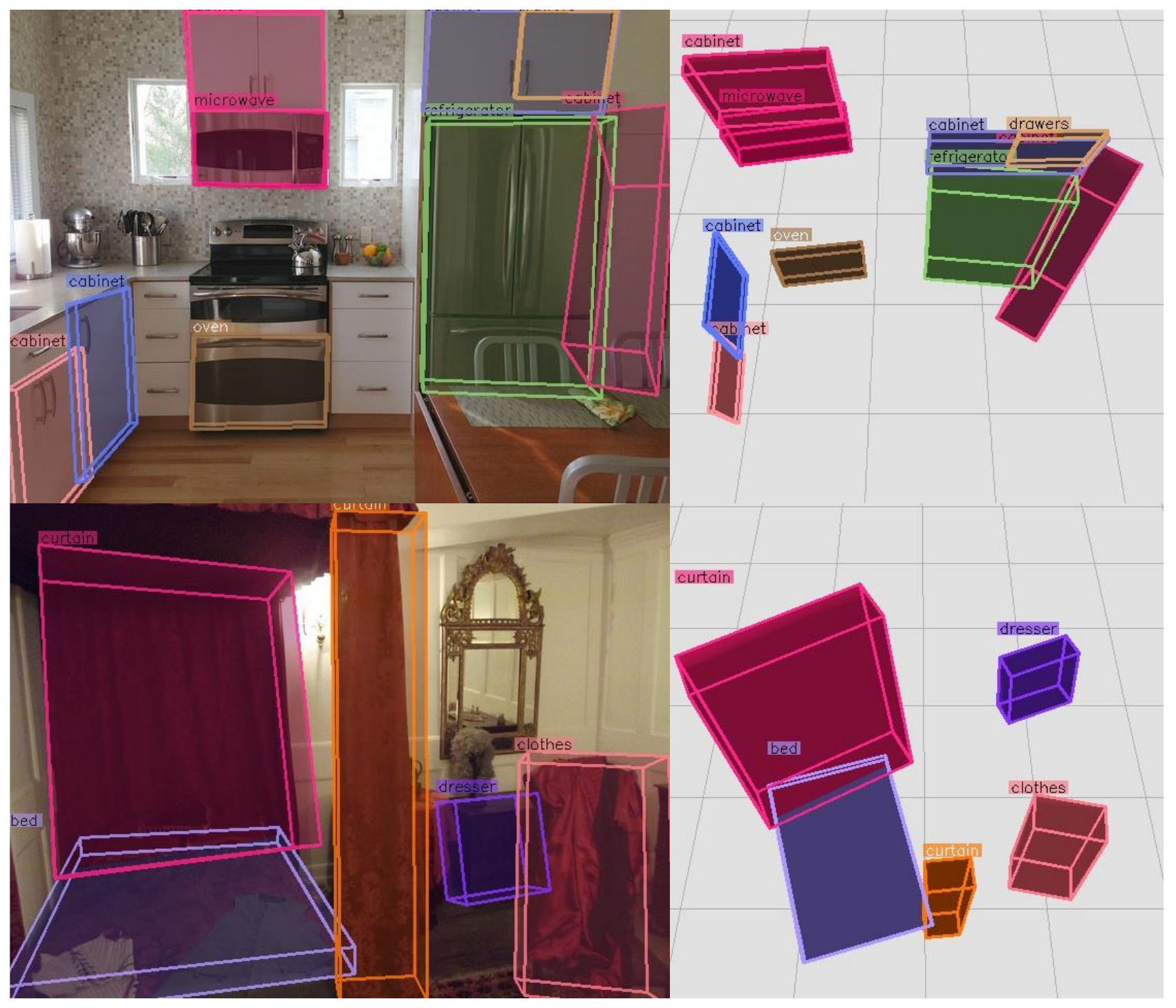}
    \caption{\textbf{Qualitative results of the 3D pseudo bounding boxes lifted from the LVIS\cite{gupta2019lvis} 2D dataset.} 
    The proposed 3D data generation method can produce accurate 3D bounding boxes in terms of categories, locations, and orientations.
    }
    \label{fig:lvis_pseudo_box}
\end{figure}

\mparagraph{Pseudo 3D Training Data.}
To quantitatively evaluate the quality of our generated pseudo 3D data, we compute the precision and recall on ScanNet200\cite{rozenberszki2022scannet200}, a 3D object detection benchmark dataset offering high-quality instance masks and ground truth 3D bounding boxes.
We meticulously choose ScanNet200 dataset because it encompasses a diverse set of 200 object categories, allowing us to validate the accuracy of the resulting pseudo 3D boxes across as many classes as possible.
%
%
As discussed in \Cref{method:3d-data-generation}, the 3D lifted point cloud, in the absence of point cloud denoising, may include numerous outlier points, typically located at the object boundaries, as illustrated in the raw point cloud shown in \Cref{fig:3d-data-generation}.
These noisy point clouds result from inaccuracies in depth map estimation and camera intrinsic parameter prediction.
As demonstrated in \Cref{table:pseudo-3d-box}, removing outliers is essential for ensuring the precision and recall of pseudo 3D boxes.
Moreover, the proposed point thresholding method is also crucial for enhancing pseudo box precision.
It effectively filters out implausible pseudo boxes with insufficient points, which may result from distant instances, heavily occluded objects, or objects with sparse and noisy points after denoising due to inaccurate depth maps and camera intrinsic parameters.
We emphasize that the precision of the pseudo 3D box is more critical than its recall for training {\ours}, as poor training examples can significantly hinder the final detection accuracy.
We show the qualitative results of the lifted 3D pseudo bounding boxes in \Cref{fig:lvis_pseudo_box}, demonstrating the high accuracy of our 3D data generation approach. 

\begin{table}[!htb]
\small
\centering
\caption{
  \textbf{Quantitaive result of the pseudo 3D box generation}.
  We report performance on ScanNet200, which offers 3D bounding box annotations for \emph{a diverse set of 200 object classes}.
  Each proposed component significantly enhances precision, which is crucial for 3D detector training.
}


\begin{tabular}{cc|cc}
\toprule
Outlier Removal & Point Thres. & Precision ($\uparrow$) & Recall ($\uparrow$)  \\ \midrule
\checkmark      & \checkmark         & \textbf{29.72}         & 16.53   \\
\checkmark      & -                  & 23.94                  & \textbf{16.85}   \\
-               & -                  & 19.75                  & 14.15   \\
\bottomrule
\end{tabular}

\label{table:pseudo-3d-box}
\end{table}

\begin{table*}[!htb]
\centering
\caption{
  \textbf{Benchmarking detection performance on Omni3D\textsubscript{IN}}.
  {\ours} demonstrates consistent improvement over the baseline methods, Cube R-CNN and its CLIP variant, in both AP\textsubscript{2D} and AP\textsubscript{3D}.
  As expected, the baseline methods fail to detect new classes in both 2D and 3D.
  In contrast, \emph{without relying on human-annotated 3D training data for new classes}, {\ours} can fairly detect new-class objects.
  We provide the relative performance ratio between our method and the original Cube R-CNN baseline in the last row.
}

\begin{tabular}{l|ccc|ccc}
\toprule
                         & $\text{AP}_\text{2D}$ & $\text{AP}_\text{2D}^\text{original}$ & $\text{AP}_\text{2D}^\text{new}$ & $\text{AP}_\text{3D}$ & $\text{AP}_\text{3D}^\text{original}$ & $\text{AP}_\text{3D}^\text{new}$ \\ \midrule
Cube R-CNN~\cite{omni3d} & 9.64 & 19.28 & 0.00 & 7.52 & \textbf{15.04} & 0.00 \\
Cube R-CNN + CLIP        & 9.18 & 18.33 & 0.02 & 7.44 & 14.88 & 0.00 \\
{\ours} (ours)                     & \textbf{12.98} {\footnotesize\textcolor{green}{($+34.6\%$)}} & \textbf{20.07} {\footnotesize\textcolor{green}{($+4.1\%$)}} & \textbf{5.89} {\footnotesize\textcolor{green}{($+\infty$)}} & \textbf{8.52} {\footnotesize\textcolor{green}{($+13.3\%$)}} & 14.39 {\footnotesize\textcolor{red}{(${-4.3\%}$)}} & \textbf{2.66} {\footnotesize\textcolor{green}{($+\infty$)}} \\
\bottomrule
\end{tabular}


\label{table:main_exp}
\end{table*}

\begin{table*}[!htb]
\centering
\caption{
  \textbf{Ablation study for {\ours}}.
  The proposed ambiguity loss and self-calibration loss both enhance the detection accuracy of new-class objects, resulting in an overall improvement in AP\textsubscript{2D} and AP\textsubscript{3D}.
}

\begin{tabular}{cc|ccc|ccc}
\toprule
Ambiguity Loss & Self-calibration Loss & $\text{AP}_\text{2D}$ & $\text{AP}_\text{2D}^\text{original}$ & $\text{AP}_\text{2D}^\text{new}$ & $\text{AP}_\text{3D}$ & $\text{AP}_\text{3D}^\text{original}$ & $\text{AP}_\text{3D}^\text{new}$ \\ \midrule
\checkmark     & \checkmark           & \textbf{12.98} & 20.07 & \textbf{5.89} & \textbf{8.52} & 14.39 & \textbf{2.66} \\
\checkmark     & -                    & 12.82 & 20.14 & 5.49 & 7.80 & 14.29 & 1.31 \\
-              & -                    & 11.56 & \textbf{20.21} & 2.91 & 7.87 & \textbf{14.62} & 1.12 \\
\bottomrule
\end{tabular}

\label{table:loss_ablation}
\end{table*}

\begin{figure*}[h]
    \centering
    \includegraphics[width=0.92\textwidth]{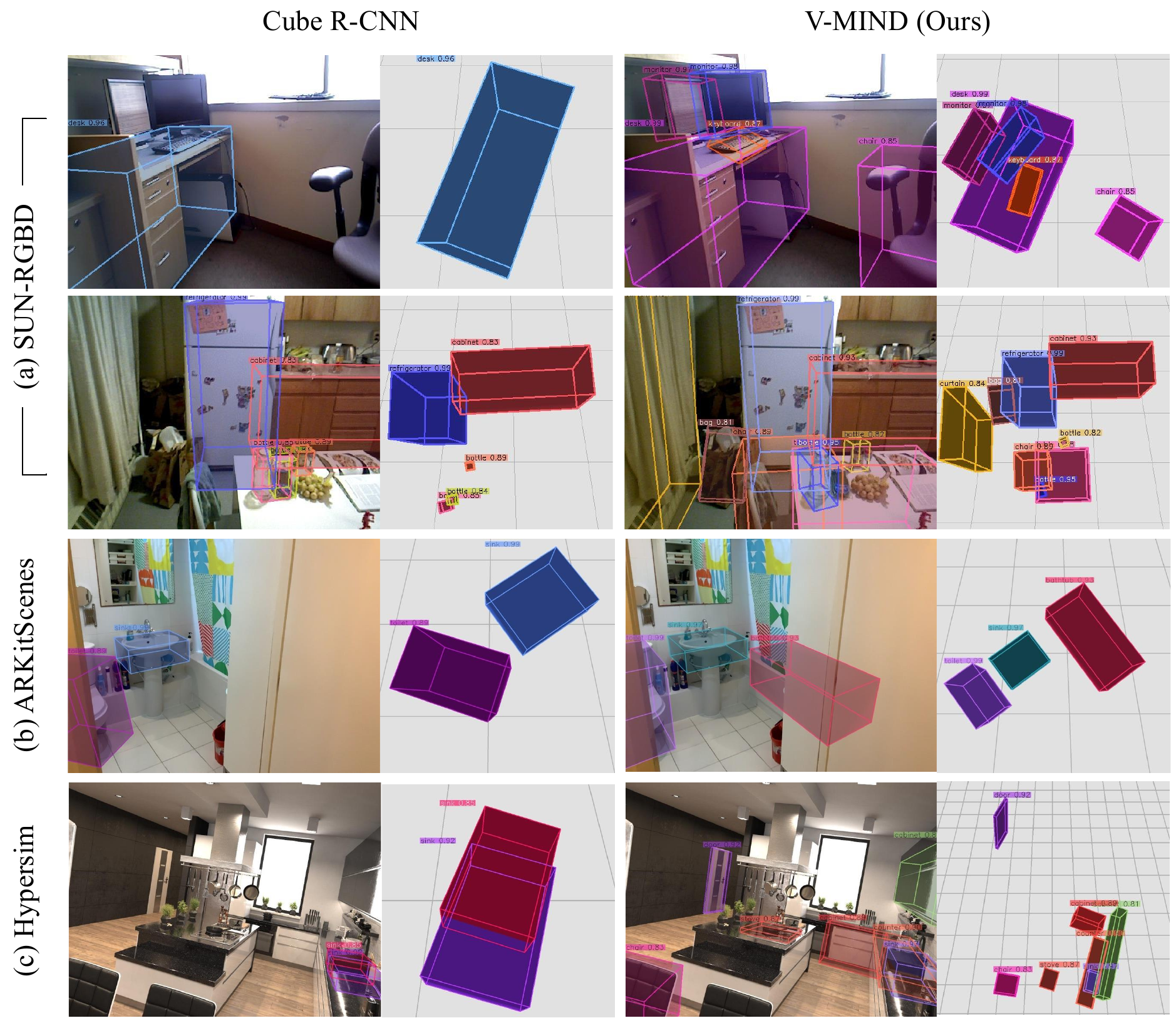}
    \caption{\textbf{Qualitative results of {\ours} and baseline Cube R-CNN on (a) SUN-RGBD, (b) ARKitScenes, and (c) Hypersim.}
    As demonstrated in (a), {\ours} effectively predicts new--class objects—specifically, a monitor and keyboard in the first row and a bag in the second row--while Cube R-CNN fails to do so.
    In (b), {\ours} successfully detects the partially occluded bathtub, whereas Cube R-CNN fails to do so.
    This demonstrates that incorporating diverse data from the augmented 2D-to-3D dataset has the potential to help address challenging detection scenarios, such as occlusion.
    %
    %
    For all examples, {\ours} shows the ability to detect a wider variety of object classes compared to the baseline Cube R-CNN.
    }
    \label{fig:qualitative}
    \vspace{-3px}
\end{figure*}

\mparagraph{Monocular 3D Detection.}
We show our detection accuracy on the $\text{Omni3D}_{\text{IN}}$ in \Cref{table:main_exp}. Apart from the original Cube R-CNN\cite{omni3d}, we add another baseline which is the Cube R-CNN with the pre-trained CLIP embeddings and train it on the original $\text{Omni3D}_{\text{IN}}$ dataset by ourselves. 
As expected, training solely with the limited classes of the original 3D dataset, the baseline methods struggle to detect new-class objects in both 2D and 3D, even with the help of the pre-trained CLIP embeddings. 
In contrast, by introducing the lifted 2D data and the proposed losses during training, {\ours} has the ability to predict new classes, 
leading to a relative improvement ratio approaching infinity ($\infty=x\,/\,0,\;\forall\,x > 0$) in $\text{AP}_\text{2D}^\text{new}$ and $\text{AP}_\text{3D}^\text{new}$. 
Although the inherent lifting errors from off-the-shelf depth and camera parameter models~\cite{bhat2023zoedepth,zhu2024tamecamera} slightly affect the 3D performance in original classes (with a relative drop of only 4.3\%), the substantial improvement in new classes still makes our method significantly outperform the original Cube R-CNN, with increases of 34.6\% in $\text{AP}_\text{2D}$ and 13.3\% in $\text{AP}_\text{3D}$.
This demonstrates the effectiveness of our proposed method in developing a versatile object detector capable of recognizing a broad range of object classes.
We show the qualitative results in \Cref{fig:qualitative}. From the visualization of predicted 3D boxes, we can clearly observe that {\ours} is able to detect a wider variety of object classes compared to the baseline model. 
Notably, these results are achieved without the need for any additional annotated 3D data, further highlighting the significance of our proposed method.

%

\subsection{Abalation Studies}
\label{sec.expAS}
We show the ablation experiment results of our proposed losses in \Cref{table:loss_ablation} and investigate the effectiveness of the pre-trained CLIP embeddings in \Cref{table:clip_ablation}.

\mparagraph{Self-calibration Loss.}
In \Cref{table:loss_ablation}, we can observe that training the model without the proposed self-calibration loss results in inferior performance in AP\textsubscript{3D}. This indicates that simply adding the 2D lifted data into training does not necessarily improve the model's performance, as it loses the ability to self-calibrate for inherent lifting errors. This issue is especially critical for new-class objects (as evidenced by the performance drop in $\text{AP}_\text{3D}^\text{new}$), since all of their training examples are derived from the lifted data.

\mparagraph{Ambiguity Loss.}
In addition to removing the 3D self-calibration loss, if we further disable another proposed ambiguity loss, a significant performance drop can be observed in both $\text{AP}_\text{2D}^\text{new}$ and $\text{AP}_\text{3D}^\text{new}$. This implies that the ambiguity issue mentioned in \Cref{method:ambiguity_loss} is indeed present and hinders the classification learning process. Specifically, when the model tries to learn to classify new classes from the lifted 2D dataset, it may simultaneously receive incorrect penalties from the original 3D dataset due to the ambiguity caused by missing annotations.
Our proposed ambiguity loss can effectively address this issue, enabling the model to successfully detect new objects, even when they lack correct 3D annotations in the original dataset.
A minor performance drop in original classes is attributed to the ambiguity loss. Without this loss, the detector only needs to classify original classes on the Omni3D dataset and treat new classes as background, which is a significantly simpler task compared to classifying both new and original classes. 




\begin{table}[!htb]
\centering
\caption{
    \textbf{Performing classification in the pre-trained vision-language space (CLIP)}, {\ours} improves the new-class and overall 3D detection performance.
}

\begin{tabular}{l|ccc}
\toprule
              & $\text{AP}_\text{3D}$ & $\text{AP}_\text{3D}^\text{original}$ & $\text{AP}_\text{3D}^\text{new}$ \\ \midrule
{\ours} w/o CLIP & 8.35 & \textbf{14.51} & 2.18 \\
{\ours} (ours)          & \textbf{8.52} & 14.39 & \textbf{2.66} \\
\bottomrule
\end{tabular}

\label{table:clip_ablation}
\end{table}
\mparagraph{Pre-trained CLIP Embeddings.}
In \Cref{table:clip_ablation}, we demonstrate that removing the pre-trained CLIP embeddings from our model architecture leads to suboptimal performance in $\text{AP}_\text{3D}$. 
This suggests that as the vocabulary size increases, it becomes more important for the model to focus on learning semantic concepts rather than merely classifying discrete categories, particularly for long-tailed new classes (as evidenced by the drop in $\text{AP}_\text{3D}^\text{new}$).
Besides, our method has the potential to further expand the number of new classes when not restricted by the available evaluation class set. We believe that this architectural design will become even more crucial as the vocabulary size continues to grow.

\section{Conclusion}
\label{section:conclusion}
By leveraging well-established monocular depth estimation techniques and camera intrinsic predictors, we generate diverse training data for 3D object detection by converting large-scale 2D images into 3D point clouds and subsequently deriving pseudo 3D bounding boxes. Through joint training with both existing 3D datasets and pseudo 3D boxes lifted from 2D datasets, combined with the proposed self-calibration loss and ambiguity loss, we develop a versatile 3D detector that achieves state-of-the-art object detection performance across a wide range of classes on the Omni3D indoor dataset.

{\small

}

\end{document}